\documentclass{article} 
\usepackage[final]{colm2026_conference}

\usepackage{microtype}
\usepackage{hyperref}
\usepackage{url}
\usepackage{booktabs}
 \usepackage{graphicx}
 \usepackage{wrapfig}
 \usepackage{adjustbox}

 \usepackage{pifont}

\usepackage{algorithm}
\usepackage{algorithmic}
\usepackage{amsmath, amssymb}
\usepackage{xcolor}
\usepackage{titletoc}
\usepackage{enumitem}
\usepackage[most]{tcolorbox}

\usepackage{lineno}

\definecolor{darkblue}{rgb}{0, 0, 0.5}
\hypersetup{colorlinks=true, citecolor=darkblue, linkcolor=darkblue, urlcolor=darkblue}

\newcommand{\ours}{Agent0}

\title{\texttt{\ours}: Unleashing Self-Evolving Agents from Zero Data via Tool-Integrated Reasoning}

\author{Peng Xia\textsuperscript{1} \quad Kaide Zeng\textsuperscript{1} \quad Jiaqi Liu\textsuperscript{1} \quad Can Qin\textsuperscript{2} \quad Fang Wu\textsuperscript{3} \quad Yiyang Zhou\textsuperscript{1}   \\ \textbf{Caiming Xiong\textsuperscript{4} \quad Huaxiu Yao\textsuperscript{1}} \\
\textsuperscript{1}UNC-Chapel Hill \quad
\textsuperscript{2}Google \quad \textsuperscript{3}Stanford University \quad 
\textsuperscript{4}Recursive Superintelligence \quad \\
\texttt{\{pxia,huaxiu\}@cs.unc.edu}
}

\begin{document}

\ifcolmsubmission
\linenumbers
\fi

\maketitle

\begin{abstract}
Large Language Model (LLM) Agents, often trained with Reinforcement Learning (RL), are constrained by a dependency on human-curated data, limiting scalability and tethering AI to human knowledge.
Existing self-evolution frameworks offer an alternative but are typically restricted by the model's inherent capabilities and single-round interactions, hindering the development of complex curricula involving tool use or dynamic reasoning.
We introduce $\texttt{\ours}$, a fully autonomous framework that evolves high-performing agents without external data through multi-step co-evolution and seamless tool integration. $\texttt{\ours}$ establishes a symbiotic competition between two agents initialized from the same base LLM: a curriculum agent that proposes increasingly challenging frontier tasks, and an executor agent that learns to solve them. We integrate external tools to enhance the executor’s problem-solving capacity; this improvement, in turn, pressures the curriculum agent to construct more complex, tool-aware tasks.
Through this iterative process, $\texttt{\ours}$ establishes a self-reinforcing cycle that continuously produces high-quality curricula.
Empirically, $\texttt{\ours}$ substantially boosts reasoning capabilities, improving the Qwen3-8B-Base model by 18\% on mathematical reasoning and 24\% on general reasoning benchmarks. 
Code is available at \href{https://github.com/aiming-lab/Agent0}{\texttt{https://github.com/aiming-lab/Agent0}}.
\end{abstract}

\section{Introduction}
Large Language Model (LLM) Agents have shown remarkable capabilities in tackling complex, long-horizon problems~\citep{qiu2025alita,qiu2025alitag,jin2025search,zhai2025agentevolver} that require extensive interaction with an environment, such as deep research~\citep{openaideepresearch,geminideepresearch} and agentic coding~\citep{jimenez2023swe,claudecode}. To optimize these complex, multi-step interactions and move beyond hard-coded workflows, Reinforcement Learning (RL) has emerged as a principal training paradigm~\citep{ouyang2022training,shao2024deepseekmath}, achieving significant progress on complex reasoning tasks.
However, the efficacy of these methods, whether Reinforcement Learning from Human Feedback (RLHF) or Reinforcement Learning from Verifiable Rewards (RLVR), relies heavily on massive, high-quality, human-curated datasets~\citep{zhang2025survey}. This dependency not only creates a severe scalability bottleneck~\citep{yue2025does}, which is time-consuming, labor-intensive, and costly, but also fundamentally tethers the potential of AI to the limits of human knowledge and annotation speed.

To break free from this reliance on human data, self-evolution frameworks have emerged as a promising alternative~\citep{zhao2025absolutezeroreinforcedselfplay,liu2025spiral,huang2025rzeroselfevolvingreasoningllm}, offering a scalable pathway by enabling models to autonomously generate their own training data. Yet, despite their potential, existing self-play or self-challenging approaches face severe constraints. First, their capabilities are capped by the model's inherent knowledge and reasoning abilities~\citep{fang2025serl,cheng2024self,zhou2025self}, causing the generated tasks to rarely surpass the model's current complexity~\citep{zhou2025evolving}, leading to learning stagnation. Second, these frameworks typically operate only in single-round interactions~\citep{li2025beyond}, failing to capture the dynamic, context-dependent nature of real-world problems. This dual limitation not only restricts the complexity of the self-generated curriculum but, more critically, hinders the model from mastering essential skills that require complex tool use or multi-step reasoning.

\begin{figure}[t]
    \centering
    \includegraphics[width=0.86\linewidth]{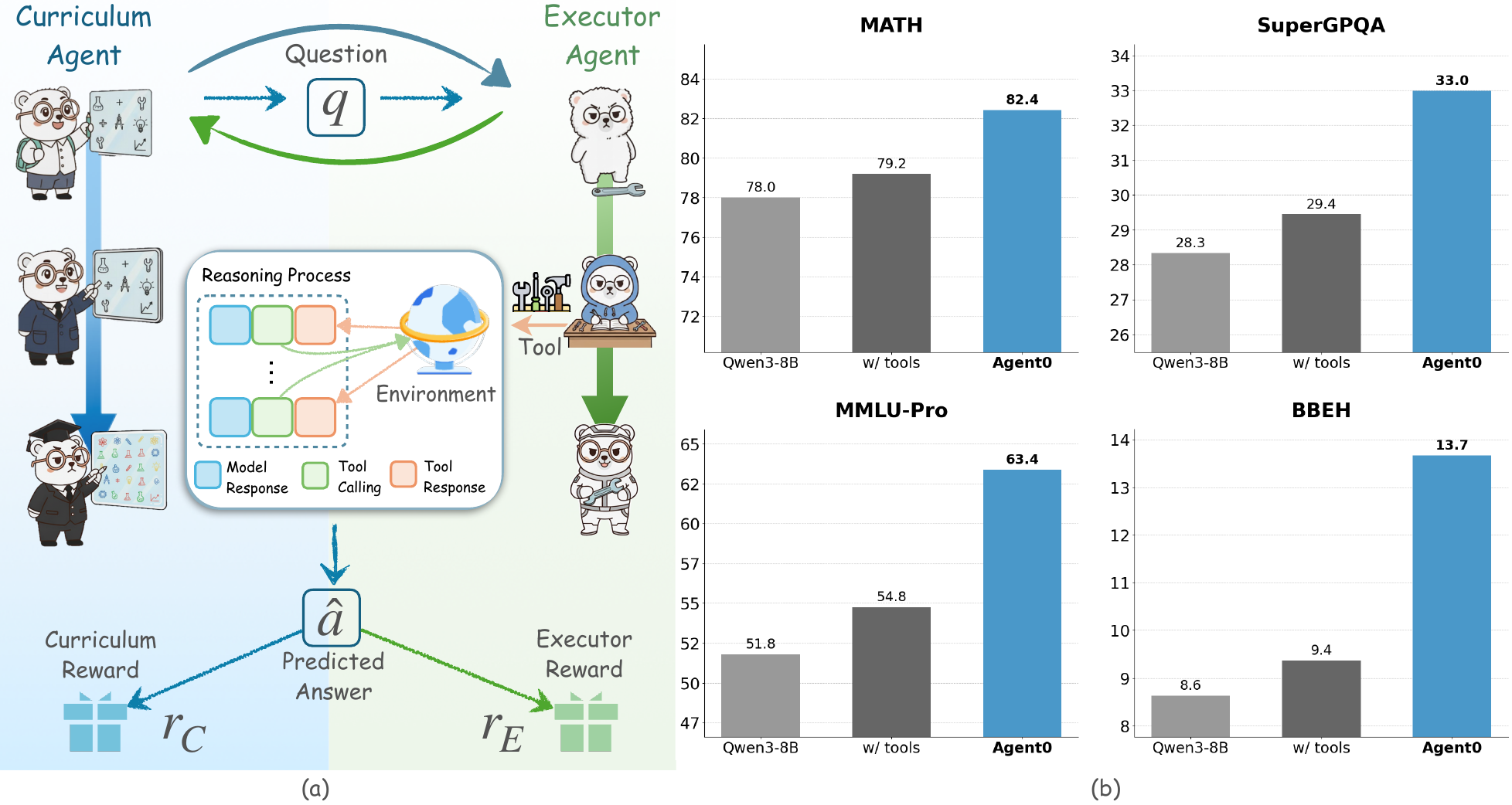} 
    \vspace{-1em}
    \caption{The $\texttt{\ours}$ autonomous co-evolution framework. The Curriculum Agent (left) uses RL to generate frontier tasks, rewarded by the Executor Agent's uncertainty and tool-use frequency. The Executor Agent (right) learn to solve them by RL. This shared tool integration drives a virtuous cycle, spiraling up task complexity and agent capability entirely from scratch.}
    \vspace{-0.5em}
    \label{fig:overview}
    \vspace{-1em}
\end{figure}

To address these challenges, as demostrated in Figure~\ref{fig:overview}, we introduce $\texttt{\ours}$, a fully autonomous framework designed to guide the evolution of agents entirely from scratch. $\texttt{\ours}$ completely eliminates the dependence on any external data or human annotations, pioneeringly combining tool integration with multi-round co-evolution. The framework's implementation begins with a base LLM from which we initialize two functionally distinct agents: \textit{an executor agent} and \textit{a curriculum agent}. These agents co-evolve through a symbiotic competition: the curriculum agent is trained using RL to propose frontier tasks that precisely challenge the executor's current capabilities, using the executor's uncertainty (\textit{i.e.,} self-consistency across multiple answers) and its frequency of tool use as reward signals. Concurrently, the executor agent is trained via RL to successfully solve these tasks, optimizing on a filtered set of challenging problems generated by the frozen curriculum agent and using pseudo-labels derived from its own majority voting. Equipping the executor with a tool enhances its problem-solving abilities, which in turn compels the tool-equipped curriculum agent to generate more complex, tool-based curricula. This establishes a virtuous cycle, driving a synchronous spiral of improvement in both agent capability and curriculum complexity. Furthermore, we extend this paradigm to support multi-turn interactions, enabling the generation of context-rich, conversational tasks that better reflect real-world problem-solving.

The primary contribution of this paper is $\texttt{\ours}$, a novel framework that autonomously evolves LLM agents from scratch through tool-augmented reasoning without relying on any external data. Across ten benchmarks spanning mathematical and general reasoning, empirical results show that $\texttt{\ours}$ achieves substantial model agnostic capability gains, improving mathematical reasoning performance by 18\% and general reasoning performance by 24\%. In addition, our analysis confirms this gain is driven by our co-evolutionary loop, where the curriculum agent learns to generate progressively complex tasks, creating a virtuous cycle of the executor's capability improvement.
\section{Preliminaries}

\label{sec:pre}

\begin{figure}[t]
    \centering
    \includegraphics[width=0.9\linewidth]{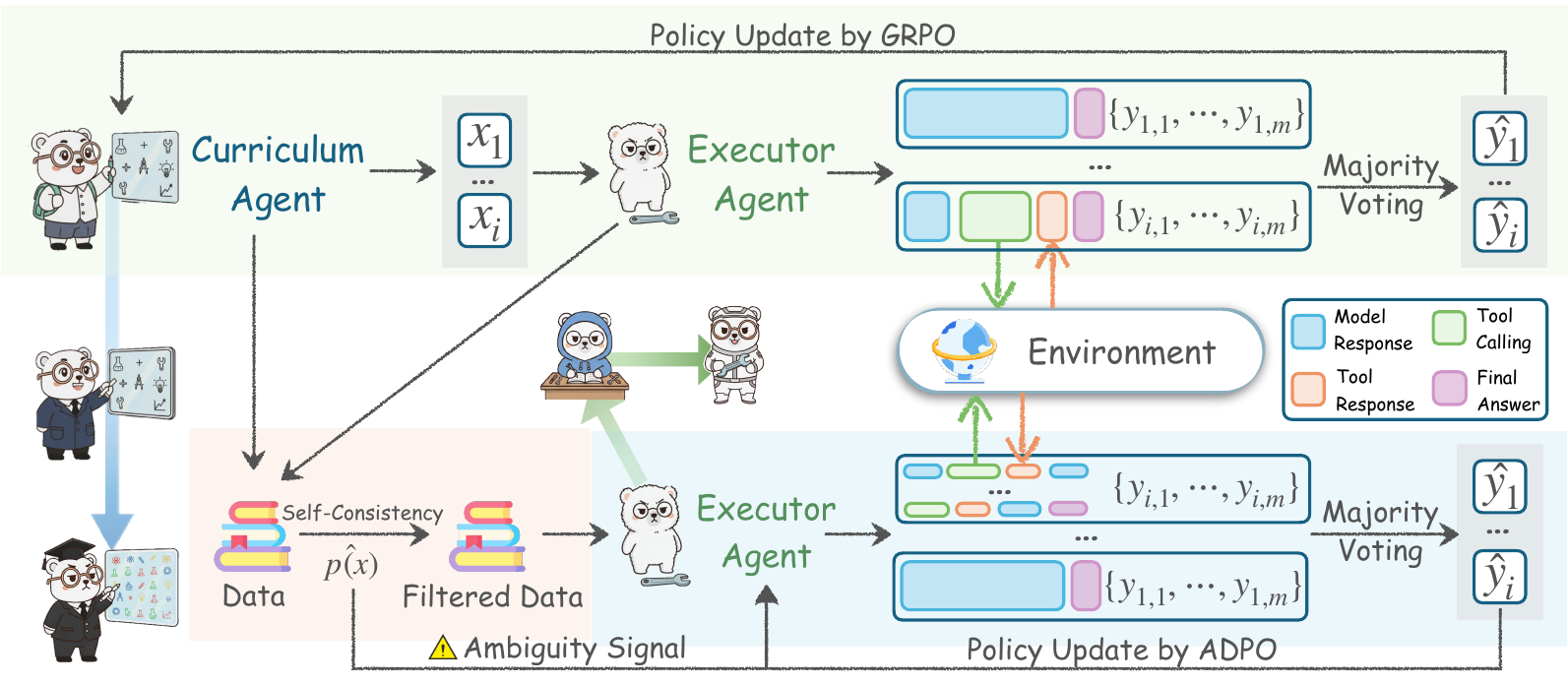}
    \caption{The $\texttt{\ours}$ co-evolutionary loop. (1) Curriculum Evolution: The Curriculum Agent $\pi_{\theta}$ is trained via RL to generate tasks, maximizing a reward $\footnotesize R_C$ based on executor Uncertainty $\footnotesize R_{\text{unc}}$, Tool Use $\footnotesize R_{\text{tool}}$ and Repetition Penalty $\footnotesize R_\text{rep}$. (2) Executor Evolution: Tasks are filtered by self-consistency score $\hat{p}$ to create a challenging dataset $\footnotesize \mathcal{D}^{(t)}$. The Executor Agent $\pi_{\phi}$ is then trained on $\footnotesize \mathcal{D}^{(t)}$ via ADPO, an ambiguity-aware RL method using majority-vote pseudo-labels $\footnotesize \tilde{y}$.}
    \vspace{-1em}
    \label{fig:method}
\end{figure}

\textbf{LLM as a Policy Agent.} We formulate the LLM as an agent, represented by a policy $\small \pi_{\theta}$ with parameters $\theta$. Given a prompt $x$, the agent autoregressively generates a response $\small y \sim \pi_{\theta}(\cdot|x)$. The general objective of reinforcement learning is to optimize $\theta$ to maximize the expected reward $\small J(\theta) = \mathbb{E}_{x \sim \mathcal{D}, y \sim \pi_{\theta}(\cdot|x)} [R(y|x)]$. 

\textbf{Group Relative Policy Optimization (GRPO)}~\citep{shao2024deepseekmath} is a reinforcement learning method that avoids training a critic by using intra-group relative rewards. For each prompt $x$, the model samples $G$ responses $\{y_1, \ldots, y_G\}$, which are scored to get rewards $\{r_1, \ldots, r_G\}$. GRPO computes normalized advantages $\hat{A}_i$ using a z-score: \begin{small}$
\hat{A}_i = \frac{r_i - \text{mean}(\{r_j\}_{j=1}^{G})}{\text{std}(\{r_j\}_{j=1}^{G}) + \epsilon_{\text{norm}}}$\end{small},
where $\epsilon_{\text{norm}}$ is a small constant for numerical stability. The policy is then updated by minimizing the following PPO-style clipped loss function~\citep{schulman2017proximal}:
\vspace{-0.3em}
\begin{equation}
\footnotesize
\label{eq:grpo}
\small  \mathcal{L}_{\text{GRPO}}(\theta) =  -\frac{1}{G} \sum_{i=1}^{G} \min\Bigg( \frac{\pi_{\theta}(x_i)}{\pi_{\theta_{\text{old}}}(x_i)} \hat{A}_i, \text{clip}\Big(\frac{\pi_{\theta}(x_i)}{\pi_{\theta_{\text{old}}}(x_i)}, 1-\epsilon, 1+\epsilon\Big) \hat{A}_i \Bigg) + \beta \text{KL}(\pi_{\theta} \| \pi_{\theta_{\text{old}}}),
\end{equation}
\vspace{-0.3em}
where $\footnotesize \frac{\pi_{\theta}(x_i)}{\pi_{\theta_{\text{old}}}(x_i)}$ is the importance sampling ratio between the current policy $\pi_{\theta}$ and the reference policy $\pi_{\theta_{\text{old}}}$ from the previous iteration. $\hat{A}_i$ is the normalized advantage, and $\epsilon$ and $\beta$ are hyperparameters. The KL-divergence term acts as a regularization penalty to stabilize training.
\section{The \texttt{\ours} Framework}
\label{sec:method}
\subsection{Framework Overview}

\texttt{\ours} is a fully autonomous, iterative co-evolutionary framework designed to enhance the capabilities of LLM agents without relying on any human-annotated data. At the core of this framework are two functionally distinct agents initialized from the same base LLM, $\pi_{\text{base}}$: (1) \textbf{Curriculum Agent ($\pi_{\theta}$)} aims to generate frontier tasks that are appropriately challenging for the current Executor Agent; (2) \textbf{Executor Agent ($\pi_{\phi}$)} aims to solve the increasingly complex tasks proposed by the Curriculum Agent.
These two agents co-evolve iteratively through a process of symbiotic competition, as illustrated in Figure~\ref{fig:method}. Each iteration $t$ of this process is divided into two stages:

\textbf{Curriculum Evolution}. We train the Curriculum Agent $\pi_{\theta}$ using RL to specialize in generating tasks that challenge the current Executor Agent $\pi_{\phi}^{(t-1)}$.

\textbf{Executor Evolution}. We use the frozen Curriculum Agent $\pi_{\theta}^{(t)}$ to generate a pool of tasks, from which we filter a challenging dataset $\mathcal{D}^{(t)}$. We then train the Executor Agent $\pi_{\phi}$ on this dataset using RL, evolving it into $\pi_{\phi}^{(t)}$.

The integration of a code interpreter tool establishes a virtuous cycle: the Executor Agent's problem-solving capabilities are enhanced by the tool, which in turn compels the tool-equipped Curriculum Agent to generate more complex, tool-based curricula. Furthermore, the framework supports multi-turn interactions, enabling the Curriculum Agent to generate context-rich, conversational tasks that better reflect real-world problem-solving.

\subsection{Curriculum Agent Training}
\label{sec:curriculum_agent}

The goal of the Curriculum Agent $\pi_{\theta}$, is to generate a prompt $x$ that maximizes a composite reward signal $R_{C}$. This reward signal is designed to quantify the challenge of task $x$ for the current Executor Agent $\pi_{\phi}$. We optimize $\pi_{\theta}$ using the GRPO algorithm.

For each task $x_i$ generated by $\pi_{\theta}$, we compute its reward by sampling $k$ responses $\{y_j\}_{j=1}^k$ from the current Executor $\pi_{\phi}$. The composite reward $R_{C}$ consists of two key components:

\noindent \textbf{Uncertainty Reward}. This reward incentivizes the Curriculum Agent to generate tasks that the Executor finds confusing or uncertain~\citep{shi2025efficient,bae2025online}. We use the Executor's self-consistency $\hat{p}(x; \pi_{\phi})$ as a proxy for uncertainty. $\hat{p}$ is defined as the proportion of the $k$ responses that vote for the majority answer ($\tilde{y}$). The reward function is designed to be maximized when $\hat{p}=0.5$, where the Executor's uncertainty is highest:
\begin{equation}
R_{\text{unc}}(x; \pi_{\phi}) = 1 - 2 \left| \hat{p}(x; \pi_{\phi}) - 0.5 \right|.
\end{equation}
This function penalizes tasks that are either too easy ($\hat{p} \to 1$) or too hard ($\hat{p} \to 0$).

\noindent \textbf{Tool Use Reward}. To drive the virtuous cycle, we must explicitly reward tasks that prompt the Executor to use its tool. We define $R_{\text{tool}}$ based on the number of tool invocations, identified by the tool response marker, i.e., \texttt{```output}, within a complete prediction $y=\pi_{\phi}(x)$. Let $N_{\text{tool}}(y)$ be the total count of these markers in $y$. The reward is then calculated as a weighted, capped value:
\begin{equation}
R_{\text{tool}}(x; \pi_{\phi}) = \gamma \cdot \min(N_{\text{tool}}(y), C)
\end{equation}
where $\gamma$ is a scaling hyperparameter for reward score and $C$ is a cap on the number of rewarded calls to prevent rewarding excessive or spurious tool use.

\noindent \textbf{Repetition Penalty}. To encourage diversity within a training batch $X$, following~\citep{huang2025rzeroselfevolvingreasoningllm}, we introduce a repetition penalty $R_{\text{rep}}$. We first compute pairwise distances between generated tasks using a similarity metric, such as BLEU score~\citep{papineni2002bleu}: $d_{ij} = 1 - \text{BLEU}(x_i, x_j)$. Tasks are then grouped into clusters $\mathcal{C} = \{C_1, ..., C_K\}$ where $d_{ij} < \tau_{\text{BLEU}}$. The penalty for a task $x_i$ belonging to cluster $C_k$ is proportional to its relative cluster size:
\begin{equation}
\small
R_{\text{rep}}(x_i) = \lambda_{\text{rep}} \frac{|C_k|}{B},
\end{equation}
where $B$ is the batch size and $\lambda_{\text{rep}}$ is a scaling factor.

\noindent \textbf{Composite Reward}. The final reward combines these signals, subtracting the repetition penalty, and is gated by a format check $R_{\text{format}}$.
\begin{equation}
\label{eq:curr}
\scriptsize
R_{C}(x_i) ={} R_{\text{format}}(x_i) \cdot \max(0, ( \lambda_{\text{unc}} R_{\text{unc}} + \lambda_{\text{tool}} R_{\text{tool}} ) - R_{\text{rep}}(x_i)),
\end{equation}
where $\lambda_{\text{unc}}$, $\lambda_{\text{tool}}$, and $\lambda_{\text{rep}}$ are hyperparameters. We use this $R_{C}$ as the reward $r_i$ in the GRPO loss.

\subsection{Executor Agent Training}
\label{sec:executor_agent}

The Executor Agent $\pi_{\phi}$'s objective is to maximize its success rate in solving tasks generated by the Curriculum Agent $\pi_{\theta}$. This stage of training is also based on GRPO.

\subsubsection{Dataset Curation and Trajectory Generation}

\noindent \textbf{Challenging Dataset Construction}.
After the Curriculum Agent $\pi_{\theta}^{(t)}$ is trained, we freeze it. We use it to generate a large pool of candidate tasks $X_{\text{pool}}$. For each task $x$ in this pool, we have the current Executor $\pi_{\phi}^{(t-1)}$ sample $k$ responses and calculate its self-consistency $\hat{p}(x)$. It is calculated as the proportion of responses that voted for this majority answer $\tilde{y}$:
\begin{equation} 
\label{eq:chall}
\small
\hat{p}(x) = \frac{1}{k} \sum_{i=1}^k \mathbb{I}(o_i = \tilde{y}), \quad \tilde{y} = \underset{y}{\text{argmax}} \sum_{i=1}^k \mathbb{I}(o_i = y),
\end{equation}
where $\mathbb{I}$ is the indicator function.
To build an efficient training curriculum, we filter for tasks that lie at the capability frontier. So we retain only those tasks whose self-consistency scores fall within an informative band:
\begin{equation}
\label{eq:data}
\scriptsize
\mathcal{D}^{(t)} = \left\{ x \in X_{\text{pool}} \mid \left| \hat{p}(x; \pi_{\phi}^{(t-1)}) - 0.5 \right| \le \delta \right\},
\end{equation}
where $\delta$ is a threshold controlling the curriculum difficulty. This filtering step ensures that $\pi_{\phi}$ trains only on tasks that are neither too easy nor too hard for it.

\noindent \textbf{Multi-Turn Rollout}. We replace the standard single-turn generation with a multi-step, tool-integrated rollout process. During this process, each of the $k$ trajectories is generated by having the policy $\pi_{\phi}^{(t-1)}$ first produce text reasoning $t_1$. When the policy emits a tool-call trigger (i.e., \texttt{```python...```} tags), generation is paused. The code $c_1$ is then executed in a sandbox, which returns an execution result or error $f_1$. This feedback $f_1$, prepended with a simple prefix like \texttt{```output...```}, is fed back to the policy. The policy then continues generating, conditioning on the history and the new feedback $[t_1 \oplus c_1 \oplus f_1 \oplus ...]$. This iterative process repeats until the policy generates a final answer $o$ (i.e., in \texttt{\{boxed{...}\}} tags), resulting in a complete, hybrid reasoning trajectory. This dynamic, interleaved feedback mechanism allows the agent to iteratively refine its reasoning and correct errors, mimicking the ``aha moment'' of self-correction.

\noindent \textbf{Pseudo-Label Advantage}. After generating $k$ full trajectories and identifying their $k$ final answers $\{o_i\}_{i=1}^k$, we use the previously determined majority answer $\tilde{y}$ as the pseudo-label. We then assign a terminal reward $\small R_i = \mathbb{I}(o_i = \tilde{y})$ to each trajectory based on whether its answer $o_i$ matches this pseudo-label.
This outcome reward $R_i$ is used to compute the advantage $A_i$ for the entire multi-step trajectory $i$.

\subsubsection{Ambiguity-Dynamic Policy Optimization}
Standard GRPO treats all training samples equally~\citep{schulman2017proximal,shao2024deepseekmath}. However, in our self-evolutionary setting, we rely on majority voting to derive pseudo-labels, which introduces two critical issues: label noise and restricted exploration on ambiguous tasks. To address these, we propose Ambiguity-Dynamic Policy Optimization (ADPO), which incorporates two key modifications motivated by the data's ambiguity signal $\hat{p}(x)$.

\begin{wrapfigure}{r}{0.33\textwidth}
    \centering
    \includegraphics[width=\linewidth]{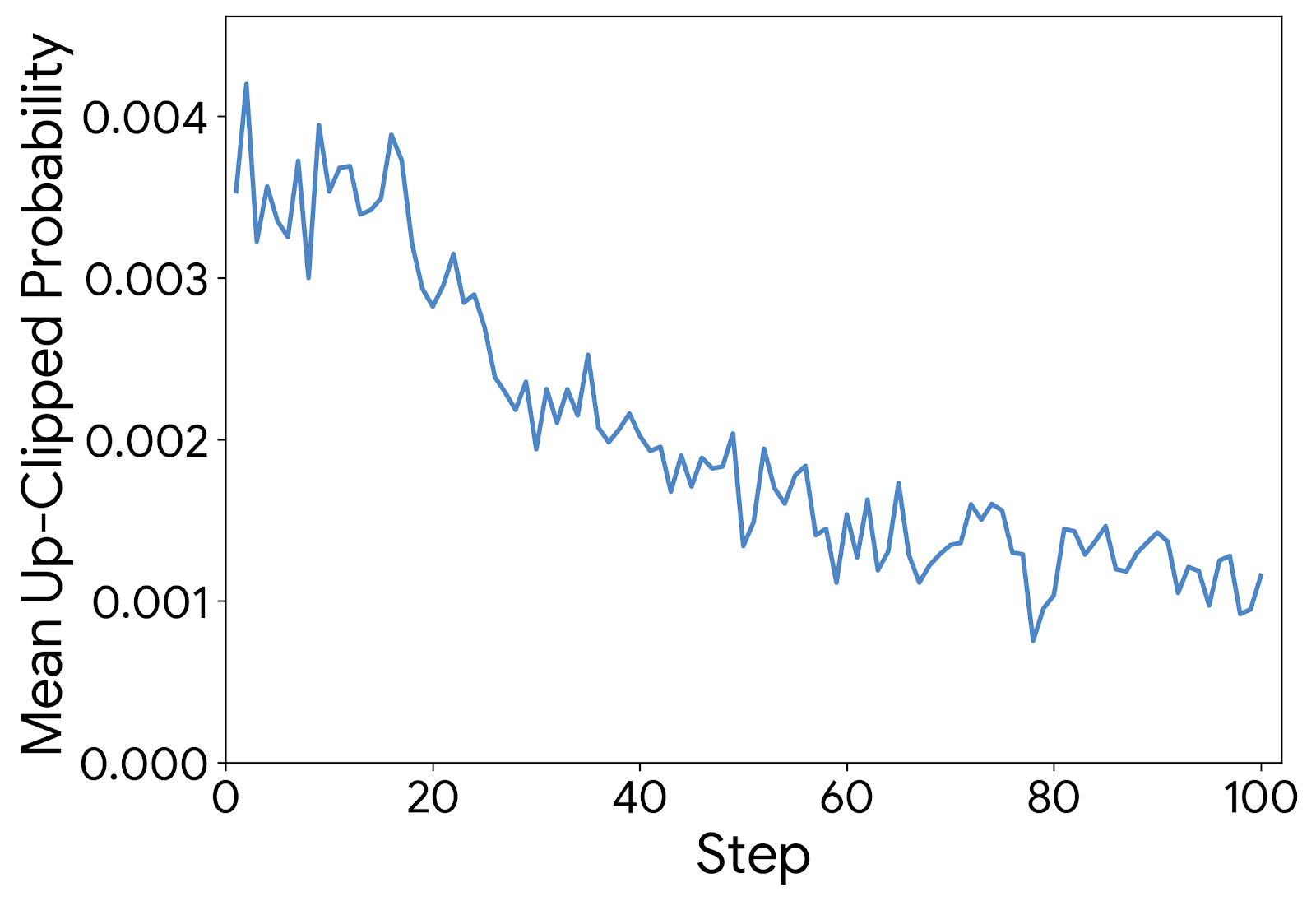}
    \vspace{-2em}
    \caption{Up-clipped token probabilities. Most up-clipped tokens have low probabilities, implying standard clipping limits exploration.}
    \vspace{-2.3em}
    \label{fig:up_clip_prob}
\end{wrapfigure}

\noindent \textbf{Ambiguity-Aware Advantage Scaling}.
The first issue is that for high-ambiguity tasks (low $\hat{p}(x)$), the majority answer is prone to errors. Directly optimizing on these noisy labels using standard GRPO risks reinforcing incorrect reasoning. To prevent overfitting to potentially inaccurate pseudo-labels, we scale the normalized advantage $\hat{A}_i$. We define a scaling factor $s(x) = f(\hat{p}(x))$, where $f$ is an increasing function of self-consistency. The advantage is modified as $\small \tilde{A}_{i}(x) = \hat{A}_{i} \cdot s(x)$. This proportionally down-weights the training signal from unreliable, low-consistency samples.

\noindent \textbf{Ambiguity-Modulated Trust Regions}.
The second issue pertains to the rigid constraints imposed by standard proximal algorithms~\citep{yu2025dapo}. While static clipping (e.g., $\epsilon$) is designed to ensure stability, it creates an asymmetric barrier to learning. As illustrated in Figure~\ref{fig:up_clip_prob}, empirical analysis reveals that the upper clipping bound is predominantly triggered by tokens with low probabilities. This indicates that the standard mechanism disproportionately ``clamps'' the growth of unlikely tokens, effectively stifling the emergence of new reasoning paths.
This restriction is particularly detrimental for high-ambiguity tasks (low $\hat{p}(x)$), where the correct reasoning often resides in the tail of the current policy distribution and requires significant updates to surface. To address this bottleneck, ADPO dynamically modulates the trust region. We define the upper clipping bound $\epsilon_{\text{high}}(x)$ as a decreasing function of $\hat{p}(x)$. This effectively relaxes the constraint for ambiguous inputs, permitting larger gradient steps to uplift potential low-probability solutions, while retaining tight bounds on confident samples to preserve stability. 

The Executor Agent is updated by minimizing the ADPO objective:
\begin{equation} \label{eq:adpo}
\small
\mathcal{L}_{\text{ADPO}}(\theta) = \mathbb{E}_{x \sim D^{(t)}} \Bigg[ -\frac{1}{G} \sum_{i=1}^{G} \min\Bigg( r_i(\theta) \textcolor{blue}{\tilde{A}_i(x)},  \text{clip}\Big(r_i(\theta), 1-\epsilon_{\text{low}}, 1+\textcolor{blue}{\epsilon_{\text{high}}(x)}\Big) \textcolor{blue}{\tilde{A}_i(x)} \Bigg)\Bigg], 
\end{equation}
where $r_i(\theta)$ is the importance sampling ratio, $\tilde{A}_i(x)$ is the ambiguity-scaled advantage, and $\epsilon_{\text{high}}(x)$ is the dynamic upper bound inversely related to $\hat{p}(x)$.
\section{Experiments}
In this section, we evaluate the performance of \ours, aiming to answer the following questions: (1) How does the performance of \ours\ compare against state-of-the-art self-evolving baselines? (2) Is the proposed co-evolutionary loop effective at progressively improving the agents' performance over multiple iterations? (3) How effective is each key component of our framework? (4) Can the mathematical reasoning abilities cultivated by \ours\ generalize to improve performance on general-domain reasoning tasks?

\subsection{Experimental Setup}
\textbf{Implementation Details}.
Our framework \texttt{\ours}, is implemented based on the VeRL~\citep{sheng2025hybridflow}. We evaluate \texttt{\ours} on two base models: Qwen3-4B-Base and Qwen3-8B-Base~\citep{yang2025qwen3}. Both the two Agent are initialized from these base models. During the co-evolutionary loop, for each task $x_i$, we sample $k=10$ responses from the Executor to compute uncertainty and generate pseudo-labels. The task filtering threshold is set to $\delta=0.25$, retaining tasks with a self-consistency $\hat{p}(x)$ between 0.3 and 0.8. For the Curriculum Agent, we set the tool reward scaling $\lambda_{\text{tool}}=0.6$ and cap $C=4$. For the Executor Agent, we integrate a sandboxed code interpreter~\citep{cheng2025fullstack} based on VeRL-Tool~\citep{jiang2025verltool}, allowing it to execute code snippets enclosed in \texttt{```python...```} tags and receive the \texttt{```output...```}.

\textbf{Baseline Methods}.
We compare \texttt{\ours} against several state-of-the-art self-improvement methods. 1) \textit{Base Model}: The pre-trained base model without any fine-tuning. 2) \textit{Base Model w/ tool}: The base model evaluated in a zero-shot setting, but given access to the code interpreter. 3) \textit{Self-Evolving Methods}: R-Zero~\citep{huang2025rzeroselfevolvingreasoningllm}, Absolute Zero~\citep{zhao2025absolutezeroreinforcedselfplay}, SPIRAL~\citep{liu2025spiral} and Socratic-Zero~\citep{wang2025socratic}.

\textbf{Evaluation Datasets and Metrics}.
\texttt{\ours} requires no human-annotated data for training. We evaluate all methods on two suites of benchmarks: 1) \textit{Mathematical Reasoning}: We use a comprehensive set including AMC, Minerva~\citep{lewkowycz2022solving}, MATH~\citep{hendrycks2021measuring}, GSM8K~\citep{cobbe2021training}, Olympiad-Bench~\citep{he2024olympiadbench}, AIME25, and AIME24. 2) \textit{General-Domain Reasoning}: To measure generalization, we use SuperGPQA~\citep{du2025supergpqa}, MMLU-Pro~\citep{wang2024mmlu}, and BBEH~\citep{kazemi2025big}. We report the accuracy (pass@1) based on greedy decoding across all benchmarks, except AMC and AIME benchmarks (mean@32).

\begin{table}[t]
\centering
\small
\caption{Comprehensive results on mathematical reasoning benchmarks. The peak performance achieved during each model's training process is highlighted in \textbf{bold}.}
\label{tab:math_results}
\resizebox{\textwidth}{!}{%
\begin{tabular}{lcccccccccc}
\toprule
\textbf{Model Name} & \adjustbox{valign=c, height=1.3ex, raise=0.4ex}{\includegraphics{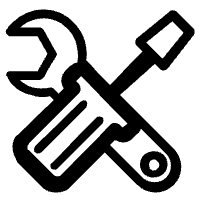}} & \adjustbox{valign=c, height=1.3ex, raise=0.4ex}{\includegraphics{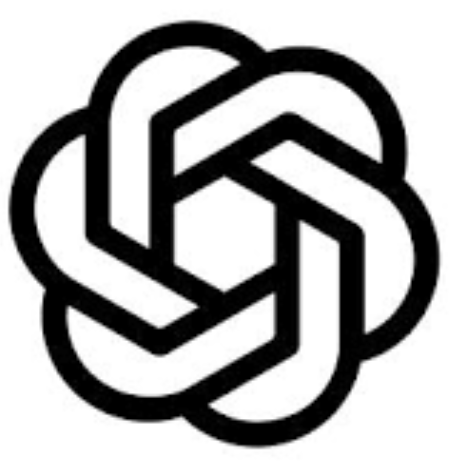}} & \textbf{AVG} & \textbf{AMC} & \textbf{Minerva} & \textbf{MATH} & \textbf{GSM8K} & \textbf{Olympiad} & \textbf{AIME25} & \textbf{AIME24} \\
\midrule
\textit{Qwen3-4B-Base} & & & & & & & & & & \\
Base Model & \textcolor{red}{\ding{55}} & \textcolor{teal}{\ding{55}}& 42.6 & 45.7 & 38.2 & 68.2 & 87.8 & 41.0 & 6.15 & 10.9 \\
Base Model \textit{w/ tool} & \textcolor{teal}{\ding{51}} & \textcolor{teal}{\ding{55}}& 44.2& 46.3 &	39.6 &	71.0 & 88.6 &	43.7& 7.71 & 12.3  \\
+ Absolute Zero & \textcolor{teal}{\ding{51}} & \textcolor{teal}{\ding{55}} & 46.4 & 50.0 & 41.9 & 76.2 & 89.3 & 41.5 & 13.4 & 12.2 \\
+ SPIRAL & \textcolor{red}{\ding{55}} & \textcolor{teal}{\ding{55}}& 47.0 & 57.5 & 42.4 & 76.4 & 91.0 & 38.4 & 10.0	& 13.3  \\
+ R-Zero & \textcolor{red}{\ding{55}} & \textcolor{teal}{\ding{55}}& 49.1 & 57.3 & 52.9 & 79.6 & 92.1 & 44.6 & 4.27 & 12.7 \\
+ $\texttt{\ours}$ &\textcolor{teal}{\ding{51}} & \textcolor{teal}{\ding{55}}& \textbf{52.5} & \textbf{60.6} & \textbf{55.6} & \textbf{80.5} & \textbf{92.6} & \textbf{46.7} & \textbf{14.1} & \textbf{17.4} \\
\midrule
\textit{Qwen3-8B-Base} & & & & & & & & & & \\
Base Model & \textcolor{red}{\ding{55}}& \textcolor{teal}{\ding{55}}& 49.2 & 52.0 & 50.0 & 78.0 & 89.1 & 44.7 & 16.7 & 13.9\\
Base Model \textit{w/ tool} & \textcolor{teal}{\ding{51}} & \textcolor{teal}{\ding{55}}& 53.2 & 60.3 & 54.9 & 79.2 & 90.7 &	47.9 &	18.7 &	20.9 \\
+ Absolute Zero & \textcolor{teal}{\ding{51}} & \textcolor{teal}{\ding{55}} & 52.6 & 62.5 & 52.9 & 76.6 & 92.0 & 47.8 & 18.2 & 18.4  \\
+ R-Zero &\textcolor{red}{\ding{55}} & \textcolor{teal}{\ding{55}}&  54.7 & 61.7 & 60.7 & 82.0 & 94.1 & 48.9 & 19.2 & 16.4 \\
+ Socratic-Zero &\textcolor{red}{\ding{55}} & \textcolor{red}{\ding{51}}& 56.1 & \textbf{63.7} & 52.4 & 81.2 & 87.3 & \textbf{55.1} &24.5 & \textbf{28.4} \\
+ $\texttt{\ours}$ &\textcolor{teal}{\ding{51}}& \textcolor{teal}{\ding{55}} & \textbf{58.2} & 62.4 & \textbf{61.3} & \textbf{82.4} & \textbf{94.5} & 54.0 & \textbf{24.8} & 28.0 \\
\bottomrule
\end{tabular}%
}
\vspace{-1em}
\end{table}

\begin{table}[t]
\centering
\small
\caption{Results on general-domain reasoning benchmarks. }
\label{tab:general_results}
\resizebox{\textwidth}{!}{%
\begin{tabular}{lccccccc}
\toprule
\textbf{Model Name} & \adjustbox{valign=c, height=1.3ex, raise=0.4ex}{\includegraphics{images/tool.png}} & \adjustbox{valign=c, height=1.3ex, raise=0.4ex}{\includegraphics{images/openai.png}} & \textbf{Overall AVG} & \textbf{MATH AVG} & \textbf{SuperGPQA} & \textbf{MMLU-Pro} & \textbf{BBEH} \\
\midrule
\textit{Qwen3-4B-Base} & \\
Base Model & \textcolor{red}{\ding{55}} & \textcolor{teal}{\ding{55}}& 27.1 & 42.6 & 20.9 &	37.4 & 7.57 \\
Base Model \textit{w/ tool} & \textcolor{teal}{\ding{51}} & \textcolor{teal}{\ding{55}}& 30.3 & 44.2 & 25.8 &	42.9 &	8.32\\
+ Absolute Zero & \textcolor{teal}{\ding{51}} & \textcolor{teal}{\ding{55}} & 33.6 & 46.4 & 27.1 & 52.6 & 8.3 \\
+ SPIRAL & \textcolor{red}{\ding{55}} & \textcolor{teal}{\ding{55}}&34.2 & 47.0& 27.1 &	53.2 & 9.57\\
+ R-Zero & \textcolor{red}{\ding{55}} & \textcolor{teal}{\ding{55}}& 34.6 & 49.1 & 27.6 &	51.5 &	10.4  \\
+ $\texttt{\ours}$  & \textcolor{teal}{\ding{51}} & \textcolor{teal}{\ding{55}} & \textbf{37.6} & \textbf{52.5} & \textbf{29.9} & \textbf{55.9} & \textbf{12.0} \\ 
\midrule
\textit{Qwen3-8B-Base} & \\
Base Model & \textcolor{red}{\ding{55}} & \textcolor{teal}{\ding{55}}& 34.5 & 49.2 & 28.3 & 51.8 & 8.6 \\
Base Model \textit{w/ tool} & \textcolor{teal}{\ding{51}} & \textcolor{teal}{\ding{55}}& 36.7 & 53.2& 29.5 & 54.8 &	9.37  \\
+ Absolute Zero & \textcolor{teal}{\ding{51}} & \textcolor{teal}{\ding{55}} & 39.9 & 52.6 & \textbf{33.5} & 62.5 & 10.8 \\
+ R-Zero & \textcolor{red}{\ding{55}} & \textcolor{teal}{\ding{55}}& 38.7 & 54.7 & 31.4 & 58.2 & 10.6  \\
+ Socratic-Zero & \textcolor{red}{\ding{55}} & \textcolor{red}{\ding{51}}& 39.2 & 56.1 & 30.1 & 60.9 & 9.5 \\
+ $\texttt{\ours}$ & \textcolor{teal}{\ding{51}} & \textcolor{teal}{\ding{55}} & \textbf{42.1} & \textbf{58.2} & 33.0 & \textbf{63.4} & \textbf{13.7} \\
\bottomrule
\end{tabular}%
}
\vspace{-1em}
\end{table}

\subsection{Main Results}

We present the main results for mathematical reasoning in Table~\ref{tab:math_results} and for general-domain reasoning in Table~\ref{tab:general_results}. 

\textbf{Comparison with Baselines.}
It significantly outperforms all compared baseline methods in both mathematics and general-domain reasoning. On Qwen3-8B-Base, \ours\ surpasses the powerful data-free method R-Zero by 6.4\% and outperforms the self-play method Absolute Zero, which utilizes a code executor, by 10.6\%. It even exceeds Socratic-Zero by 3.7\%, which relies on external OpenAI APIs. This demonstrates the superiority of \ours's self-evolution approach. By using tools to interact with the environment, the agent effectively enhances the quality and diversity of questions generated by the curriculum agent. Similarly, for the execution agent, this more effectively improves its problem-solving capabilities. 

\textbf{Generalization to General-Domain Tasks.} Furthermore, Table~\ref{tab:general_results} shows strong evidence of generalization. On Qwen3-8B, \ours\ achieves the highest overall average score among all approaches, significantly outperforming other data-free methods. This indicates that the complex, multi-step reasoning abilities we cultivated in the execution agent by using the curriculum agent with tools, can be effectively transferred to general-domain tasks.

\subsection{Analysis}
In this section, we provide a detailed analysis of each module’s performance, along with a series of analytical experiments, to better understand the performance gains.

\textbf{Ablation Study}.
As shown in Table~\ref{tab:aba}, we conducted a series of ablation experiments to evaluate the impact of each component in our method.
Specifically, we evaluate the impact of: (1) the curriculum agent's training, (2) the tool reward, (3) the repetition penalty, (4) our ambiguity scaling mechanism, and (5) the multi-turn reasoning capability.
\begin{wraptable}{r}{0.5\linewidth}
\vspace{-1em}
\footnotesize
\centering
\caption{
  Ablation study of \ours.
}
\label{tab:ablation}
\resizebox{0.99\linewidth}{!}{
\begin{tabular}{lcc}
\toprule
Method & General AVG & Math AVG  \\
\midrule
\ours & 36.7 & 58.2  \\
\midrule
\multicolumn{3}{l}{\textit{Curriculum Agent}} \\
\quad w/o Training & 29.5 & 46.8  \\
\quad w/o Tool Reward & 31.8 & 48.7  \\
\quad w/o Repetition Penalty & 31.3 & 47.9  \\
\midrule
\multicolumn{3}{l}{\textit{Execution Agent}} \\
\quad w/o ADPO & 34.9 & 56.2  \\
\quad w/o Multi-turn & 35.3 & 55.9 \\
\bottomrule
\end{tabular}
} 
\label{tab:aba}
\vspace{-1em}
\end{wraptable}
For the curriculum agent, without training, the performance significantly drops by 9.3\%. This reflects the value of the learned curriculum. Next, when the tool reward is not included, the model's performance drops by 7.2\%. This tests our core hypothesis that explicitly rewarding tool-use tasks is necessary. It shows a severe performance degradation when we remove the diversity component, indicating that $R_{\text{rep}}$ is highly effective for curriculum diversity, particularly for general tasks.
As for the execution agent, training it using the original GRPO with standard advantage and clipping resulted in a performance drop of 1.9\%. This is because the original algorithm does not account for the reliability of pseudo-labels, demonstrating the effectiveness of our proposed ambiguity scaling mechanism. The introduction of multi-turn reasoning played a significant role in boosting \ours's performance, especially for complex mathematical reasoning that requires multi-turn reasoning.
\begin{wrapfigure}{r}{0.45\textwidth}
\vspace{-1em}
    \centering
    \includegraphics[width=0.99\linewidth]{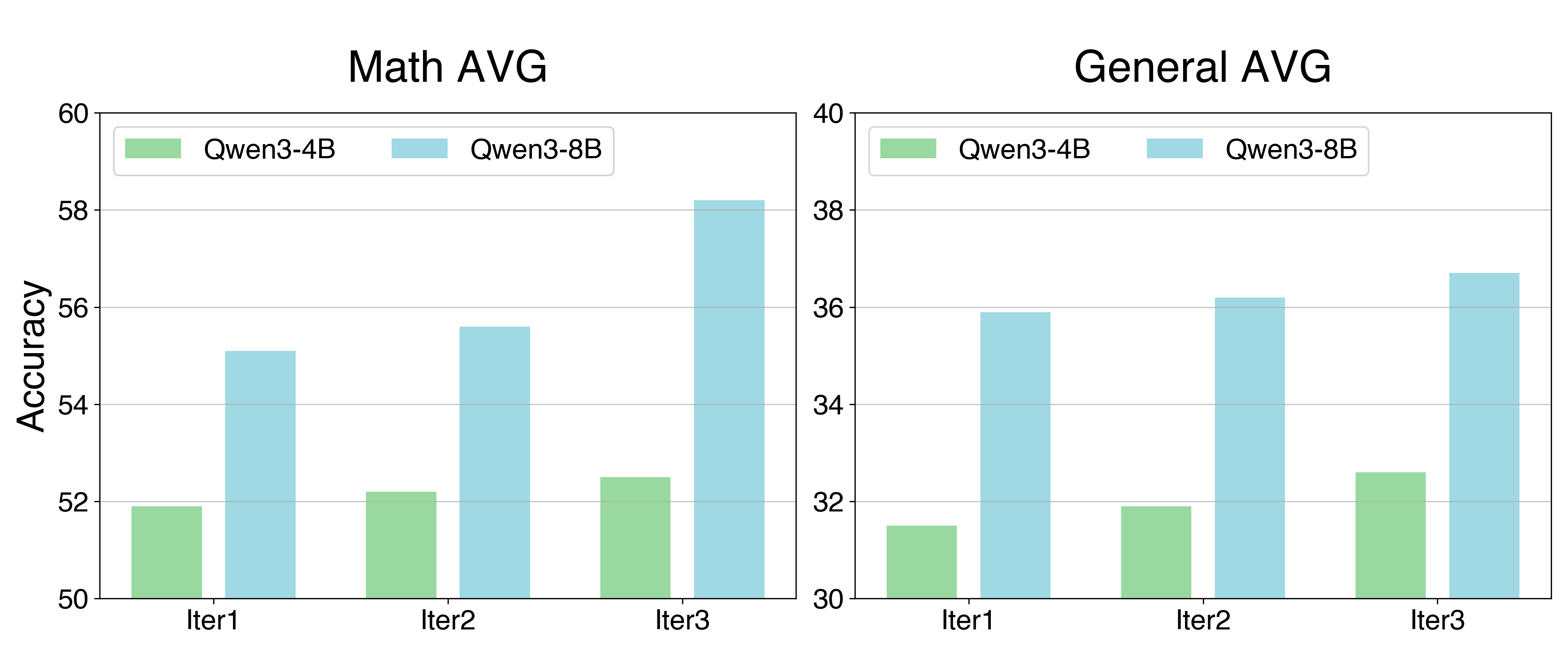} 
    \vspace{-1.5em}
    \caption{Performance on mathematical and general reasoning benchmarks, showing consistent improvement for both Qwen3-4B and Qwen3-8B across three co-evolutionary iterations.}
    \vspace{-1em}
    \label{fig:evo}
\end{wrapfigure}

\textbf{Consistent Improvement through Co-Evolution.}
As shown in Figure~\ref{fig:evo}, our method demonstrates stable and progressive improvement during the iterative process. On Qwen3-8B-Base, the average math score improved from 55.1 (Iter 1) to 56.5 (Iter 2), peaking at 58.2 (Iter 3). In addition to mathematics, \ours\ showed the similar trend on other general-domain reasoning tasks, with an average improvement of 2\% per iteration compared to the previous one. This iterative gain validates the effectiveness of our co-evolutionary loop. With the involvement of tools, the curriculum agent progressively generates more difficult tasks, while the execution agent learns to solve these tasks more efficiently. 

\begin{wraptable}{r}{0.34\linewidth}
\vspace{-2em}
    \scriptsize
    \centering
    \caption{Comparison on non-tool and other tool-integrated baselines.}
    \vspace{0.2em}
    \begin{tabular}{lcc}
    \toprule
    Model & MATH & General \\
    \midrule
    Qwen3-4B & 42.6 & 22.0 \\
    \colorbox{gray!10}{\textit{w/o Tool}} \\
    +SPIRAL & 47.0 & 30.0 \\
    +R-Zero & 49.1 & 29.8 \\
    \colorbox{gray!10}{\textit{w/ Tool}} \\
    +TIR & 44.2 & 25.7 \\
    +Absolute Zero & 46.4 & 29.3 \\
    +\texttt{\ours} & \textbf{52.5} & \textbf{32.6} \\
    \bottomrule
    \end{tabular}
    \label{tab:tool}
    \vspace{-2em}
\end{wraptable}
\textbf{Strategic Tool Integration Matters.} 
Our advantage lies not just in having a tool, but in learning how to use it. As shown in Table~\ref{tab:tool}, merely providing a tool (i.e., Base Model w/ Tool) yields a slight performance boost. \ours\ significantly outperforms other tool-using baselines, such as Absolute Zero. \ours\ also significantly surpasses non-tool methods like R-Zero and SPIRAL. This indicates that our curriculum agent, by using the $R_{\text{tool}}$ reward to explicitly incentivize the generation of complex tasks requiring tool use, is far more effective than methods that only use tools for validation  or do not use tools at all. Furthermore, the execution agent utilizes the tool in conjunction with multi-step reasoning, which also leads to performance gains, resulting in co-evolution.

\textbf{Evolution of Task Difficulty and Tool Use}.
We analyze the tasks generated by the curriculum agent during the training iterations. 
As shown in Table~\ref{tab:tool_evolution}, the pass rate of the execution agent (from Iteration 1) progressively decreases when evaluated on task sets generated by the curriculum agents from Iterations 1, 2, and 3. 
\begin{wraptable}{r}{0.6\linewidth}
\vspace{-2em}
    \centering
    \caption{Evolution of Task Difficulty and Tool Use.
    }
    \label{tab:tool_evolution}
    \resizebox{0.99\linewidth}{!}{ 
    \begin{tabular}{lcc}
        \toprule
        Dataset & Pass Rate (Executor$_{\text{Iter 1}}$) & Avg. Tool Calls \\
        \midrule
        $\mathcal{D}_{\text{Iter 1}}$ & 64.0 & 1.65 \\
        $\mathcal{D}_{\text{Iter 2}}$ & 58.5 & 2.10 \\
        $\mathcal{D}_{\text{Iter 3}}$ & 51.0 & 2.60 \\
        \bottomrule
    \end{tabular}
    }
    \vspace{-1.5em}
\end{wraptable}
This indicates that the task difficulty is gradually increasing, confirming that the curriculum adapts to the improvement in the executor's capabilities. 
More importantly, the average number of tool calls per generated task steadily increases across iterations. This directly proves that our $R_{\text{tool}}$ reward successfully guides the curriculum agent to generate more complex and tool-reliant problems, thereby driving a virtuous cycle.

\begin{figure}[t]
    \centering
    \includegraphics[width=0.85\linewidth]{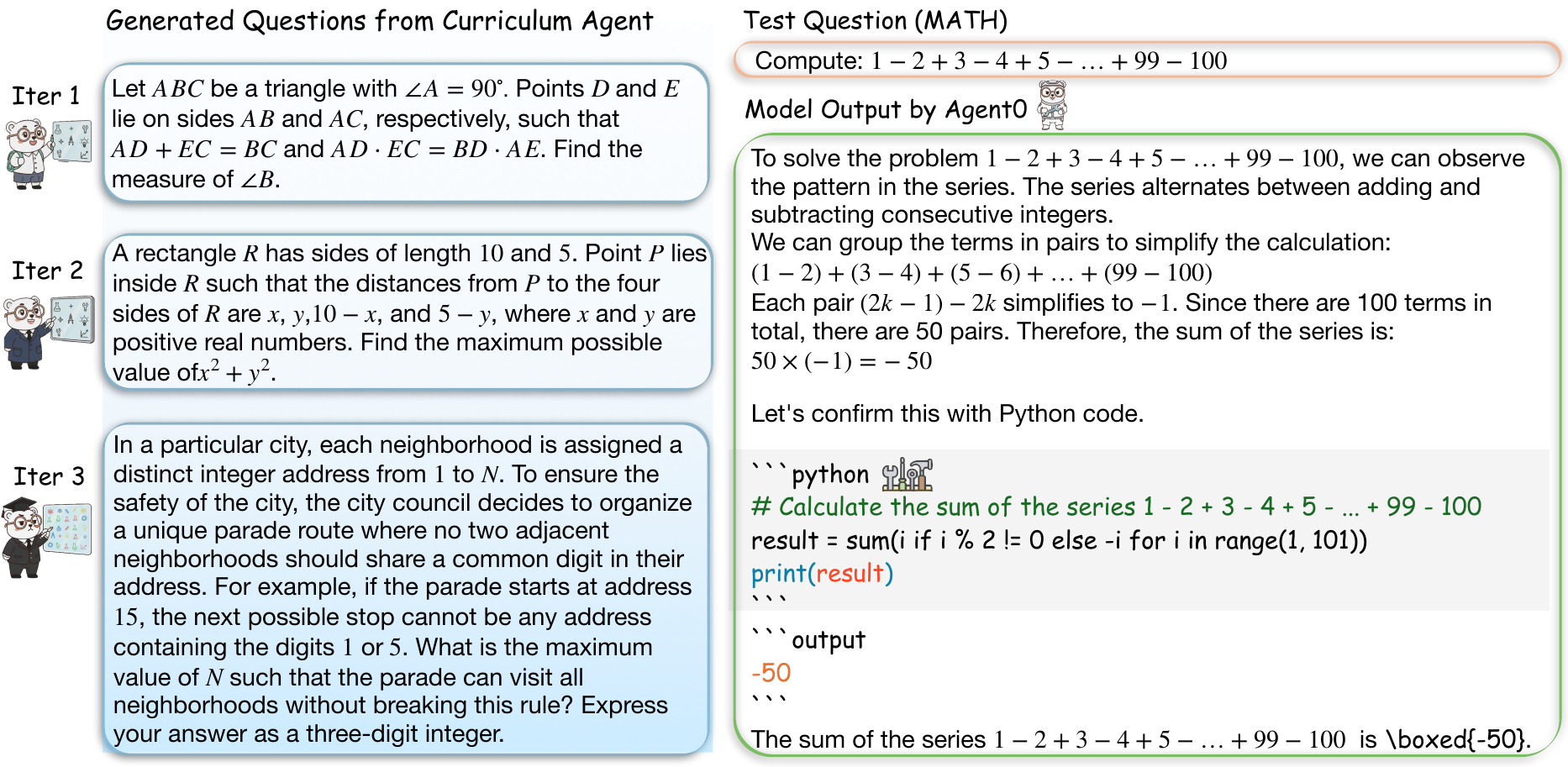} 
    \caption{Qualitative Case Analysis. 
    Left: Examples of generated questions showing a clear increase in complexity and diversity from Iter 1 to Iter 3. Right: A demonstration of \ours's solving process, utilizing a hybrid approach of mathematical reasoning and Python code execution to solve a standard MATH problem.
    }
    \vspace{-1em}
    \label{fig:case}
\end{figure}

\textbf{Qualitative Analysis}.
Figure~\ref{fig:case} illustrates the co-evolution of task complexity and solving proficiency. The Curriculum Agent effectively generates increasingly difficult problems, progressing from basic geometry (Iter 1) to complex constraint satisfaction tasks (Iter 3). Simultaneously, the Executor \ours\ demonstrates reliable problem-solving capabilities. In the provided example, the agent effectively combines natural language reasoning to identify patterns with the Python code interpreter to verify calculations, validating the model's ability to handle hybrid reasoning tasks.
\section{Related Work}
\textbf{Self-Evolving from Zero Data.} 
The paradigm of self-evolution, where LLMs generate their own training data, has gained significant traction~\citep{dong2024self,fang2025serl,yang2025spell,kuba2025language}. This approach ranges from dual-agent ``Coder-Tester'' setups in verifiable domains~\citep{lin2025learning,wang2025co} to fully autonomous frameworks~\citep{zhao2025absolutezeroreinforcedselfplay,huang2025rzeroselfevolvingreasoningllm,wang2025socratic,liu2025spice,zhang2025evolvesearch,wu2025evolver,lu2025search} that learn to generate novel problems from scratch. To guide this learning, many methods use label-free reinforcement learning, relying on heuristic reward signals such as output confidence~\citep{li2025confidence} or consistency~\citep{zhang2025consistent,prabhudesai2025maximizing,yu2025restrain}. However, these systems are critically limited by the model's inherent knowledge~\citep{han2025alignment}, causing curriculum stagnation as tasks rarely surpass the model's current complexity. $\texttt{\ours}$ breaks this cap by integrating an external tool, providing external problem-solving power. 
However, without external tools, such closed-loop systems risk mode collapse and curriculum stagnation, as they remain bounded by the model's inherent knowledge. \texttt{\ours} breaks this ceiling by integrating an external tool to introduce objective problem-solving power.

\textbf{Tool-Integrated Reasoning (TIR).}  
Applying Reinforcement Learning (RL)~\citep{jaech2024openai,bae2025online} to enhance LLM tool-use is a growing field. Many approaches rely on domain-specific data or supervised fine-tuning~\citep{jin2025search,feng2025retool, li2025torl}. The more general Zero RL setting, however, is notoriously unstable in multi-turn scenarios. Recent advances in TIR address these challenges through three key dimensions: stability, generalization, and complexity. To stabilize learning dynamics, methods like ASPO~\citep{lin2025understanding} and SimpleTIR~\citep{xue2025simpletir} introduce theoretical guarantees and gradient filtering for void turns. Beyond stability, \cite{chen2025can} demonstrate the cross-domain transferability of tool-use skills. Finally, to handle complex multi-turn scenarios, advanced techniques optimize for long-horizon planning~\citep{gao2025beyond,erdogan2025plan}, memory management~\citep{yan2025memory}, and interaction efficiency~\citep{wang2025otc}.
\section{Conclusion}
We introduce $\texttt{\ours}$, a fully autonomous framework where a curriculum agent and an executor agent co-evolve without any human-curated data. We integrated a code interpreter into the loop, which creates a virtuous cycle: the tool-equipped executor's improving capabilities drive the curriculum agent to generate progressively harder tasks. Our experiments show that $\texttt{\ours}$ significantly enhances the reasoning abilities of base LLMs. It demonstrates a scalable and effective pathway for evolving highly capable agents, breaking the dependency on human-annotated datasets.

\section*{Acknowledgments}
We thank Chengsong Huang for helpful discussions. This work is partially supported by the AI for Math Fund from Renaissance Philanthropy. The Authors also acknowledge the National Artificial Intelligence Research Resource (NAIRR) Pilot, Purdue Anvil AI for contributing to this research result.

\section*{Ethics Statement}
This paper introduces Agent0, a fully autonomous framework that evolves LLM agents without relying on human-curated data, thereby eliminating the severe scalability bottlenecks and labor costs associated with traditional annotation. While this approach significantly accelerates agent training and pushes AI beyond the limits of human knowledge, the autonomous generation and execution of code introduce potential safety risks. To mitigate these risks, our framework strictly confines all tool integration and code execution within an isolated, secure sandbox, preventing unintended interactions with external environments. Furthermore, because Agent0 relies on a self-reinforcing co-evolutionary loop without direct human oversight, careful consideration must be given to how these models develop complex reasoning. Future deployments of such self-evolving systems should continuously monitor the agent's trajectory to ensure that autonomously generated curricula align with safe and intended real-world behaviors.

\section*{Reproducibility Statement}
To ensure the full reproducibility of Agent0, we have thoroughly documented our experimental setup, hyperparameters, and system configurations. The complete source code will be open-sourced upon publication. All evaluation datasets are publicly available benchmarks detailed in Appendix~\ref{bench}. Our framework is built upon the open-source VeRL library and VeRL-Tool. The experiments are conducted using the publicly accessible Qwen3-4B-Base and Qwen3-8B-Base models for both the Curriculum and Executor agents. A comprehensive list of training hyperparameters is provided in Appendix~\ref{hyper}. The system and user prompts used for the agents and the judge, and  are fully disclosed in Appendix~\ref{prompt}. We detail the isolated Python code execution sandbox in Appendix~\ref{sandbox}. The architecture relies on a Distributed Execution Orchestrator built on Flask and vLLM to ensure safe, stable, and high-throughput multi-step reasoning evaluation.

\bibliography{colm2026_conference}
\bibliographystyle{colm2026_conference}

\newpage
\appendix
\onecolumn
\startcontents[appendix]
\printcontents[appendix]{ }{0}{\section*{Appendix}}

\section{Experimental Details}
\subsection{Hyperparameter Settings}
\label{hyper}
\textbf{Executor Agent Training}
\begin{itemize}[leftmargin=*]
    \item Global Batch Size: 128
    \item Learning Rate: $1 \times 10^{-6}$
    \item Weight Decay: $1 \times 10^{-2}$
    \item KL Penalty Coefficient ($\lambda_{KL}$): $1 \times 10^{-2}$
    \item Max Steps: 40
    \item Number of Rollouts: 16
    \item Rollout Temperature: 1.0
    \item Rollout Top-p: 0.99
\end{itemize}
\textbf{Curriculum Agent Training}
\begin{itemize}[leftmargin=*]
    \item Global Batch Size: 128
    \item Learning Rate: $1 \times 10^{-6}$
    \item Weight Decay: $1 \times 10^{-2}$
    \item KL Penalty Coefficient ($\lambda_{KL}$): $1 \times 10^{-2}$
    \item Max Steps: 5
    \item Number of Rollouts: 4
    \item Rollout Temperature: 1.0
    \item Rollout Top-p: 0.99
\end{itemize}

\subsection{Prompt}
\label{prompt}
This section presents the prompt templates used for the executor and curriculum agent, and judge prompt in Table~\ref{tab:prompt1}, Table~\ref{tab:prompt2} and Table~\ref{tab:prompt3}. 

\begin{table}[h]
\centering
\caption{Prompt template used for executor agent.}
\begin{tcolorbox}[colframe=black, colback=gray!10, coltitle=black, sharp corners]
\textbf{System Prompt:} \\
A conversation between User and Assistant. The user asks a question, and the Assistant solves it. The assistant first thinks about the reasoning process in the mind and then provides the user with the answer. User: Please integrate natural language reasoning with programs to solve the problem above. If you want to run any python code, write code in the python markdown code block and the execution will be appended in an output code block like \texttt{```python you code here```} 
\texttt{```output result here```}. Please put your final answer within \texttt{\text{\\boxed\{\}}}.

\textbf{User Prompt:} \\
\{problem\}
\end{tcolorbox}
\label{tab:prompt1}
\end{table}

\begin{table}[h]
\centering
\caption{Prompt template used for curriculum agent.}
\begin{tcolorbox}[colframe=black, colback=gray!10, coltitle=black, sharp corners]
\textbf{System Prompt:} \\
You are an expert competition-math problem setter. FIRST, in your private scratch-pad, think step-by-step to design a brand-new, non-trivial problem. The problem could come from any field of mathematics, including but not limited to algebra, geometry, number theory, combinatorics, prealgebra, probability, statistics, and calculus. Aim for a difficulty such that fewer than 30\% of advanced high-school students could solve it. Avoid re-using textbook cliches or famous contest problems. 
THEN, without revealing any of your private thoughts, output exactly the following two blocks:

\texttt{<question>}

{The full problem statement on one or more lines}

\texttt{</question>}

\texttt{\\boxed\{final answer\}}

Do NOT output anything else—no explanations, no extra markup.

\textbf{User Prompt:} \\
Generate one new, challenging reasoning question now. Remember to format the output exactly as instructed.
\end{tcolorbox}
\label{tab:prompt2}
\end{table}

\begin{table}[h]
\centering
\caption{Prompt template used for judging. We use the GPT-4o~\citep{achiam2023gpt} as the judge model (temperature: 0.1).}
\begin{tcolorbox}[colframe=black, colback=gray!10, coltitle=black, sharp corners]
\textbf{System Prompt:} \\
You are a math answer checker.

\textbf{User Prompt:} \\
Hi, there is an answer: \texttt{\{answer\}}, and the ground truth answer is: \texttt{\{response\}},
please check whether the answer is correct or not, and return the **only** Yes or No.
\end{tcolorbox}
\label{tab:prompt3}
\end{table}

\subsection{Sandbox Configuration}
\label{sandbox}
We integrated a Python-based code execution sandbox~\citep{cheng2025fullstack} to enable verification and algorithmic reasoning. The system comprises two core components: a Multi-Turn Interaction Protocol and a Distributed Execution Orchestrator, supported by a robust infrastructure.

\textbf{Multi-Turn Interaction Protocol.}
We employ a ``stop-and-go'' strategy to facilitate multi-step reasoning: 
The model is instructed to generate reasoning followed by executable Python code within markdown delimiters.
Upon detecting a code block via regex, generation halts. The code is extracted and run in an isolated sandbox, capturing stdout or stderr.
Execution results are appended to the conversation history. The model then interprets these results to derive a final answer formatted in LaTeX boxed notation ($\text{\\boxed}\{\cdots\}$).

\textbf{Distributed Execution Orchestrator.}
To manage parallel candidate generation (e.g., $N=10$), we implemented a load-balancing mechanism:
Execution is decoupled into isolated worker nodes to prevent interference with the main inference server.
A thread-safe Round-Robin scheduler distributes requests across nodes. A ThreadPoolExecutor manages asynchronous calls, preventing main loop blocking. 
The system robustly handles network timeouts and failures, feeding error messages back to the model for potential self-correction.

\textbf{Infrastructure.}
Built on Flask and vLLM~\citep{kwon2023efficient}, the system ensures high throughput. To maintain stability, a background thread performs low-priority tensor operations during idle periods, preventing GPU deep sleep and ensuring consistent latency.

\section{Overview of the Baselines}
\begin{itemize}[leftmargin=*]
    \item Base Model: The pre-trained base model without any fine-tuning.
    \item Base Model w/ tool: The base model evaluated in a zero-shot setting, but given access to the code interpreter.
    \item R-Zero~\citep{huang2025rzeroselfevolvingreasoningllm}: A self-evolving framework that operates from zero data, but does not utilize external tools.
    \item Absolute Zero~\citep{zhao2025absolutezeroreinforcedselfplay}: A self-play method that does use a code executor for verification, representing a strong tool-aware baseline. 
    \item SPIRAL~\citep{liu2025spiral}: A self-play method based on zero-sum games and multi-turn interactions.
    \item Socratic-Zero~\citep{wang2025socratic}: A strong baseline representing methods that leverage external proprietary models for reasoning assistance.
\end{itemize} 

\section{Evaluation Benchmarks}
\label{bench}
\begin{itemize}[leftmargin=*]
    \item AMC: A collection of problems from standard American middle and high school math competitions, serving as a foundational benchmark for pre-collegiate mathematical reasoning.
    \item Minerva~\citep{lewkowycz2022solving}: The dataset evaluates the model's ability to handle formal scientific notation and solve complex STEM-related questions.
    \item MATH~\citep{hendrycks2021measuring}: A comprehensive dataset of challenging high school competition problems across various subfields (e.g., algebra, geometry), requiring complex heuristic search and multi-step derivation.
    \item GSM8K~\citep{cobbe2021training}: A classic benchmark consisting of high-quality grade school math word problems that test the model's ability to perform multi-step logic using basic arithmetic operations.
    \item Olympiad-Bench~\citep{he2024olympiadbench}: An advanced benchmark aggregating extremely difficult problems from Chinese and International Mathematical Olympiads, designed to probe the upper limits of LLM reasoning capabilities.
    \item AIME24 \& AIME25: These datasets comprise problems from the 2024 and 2025 American Invitational Mathematics Examinations, serving as a rigorous test of advanced problem-solving on recent, likely uncontaminated data.
    \item SuperGPQA~\citep{du2025supergpqa}: An evolution of the GPQA dataset~\citep{rein2024gpqa}, this benchmark features difficult graduate-level questions across scientific domains that are challenging even for experts, specifically designed to minimize data contamination and retrieval shortcuts.
    \item MMLU-Pro~\citep{wang2024mmlu}: An enhanced version of the MMLU benchmark~\citep{hendrycks2020measuring} that introduces harder questions, increased options, and more complex reasoning requirements to better differentiate between top-tier language models.
    \item BBEH~\citep{kazemi2025big}: A selected subset of challenging tasks from the Big-Bench suite~\citep{srivastava2023beyond}, focusing on areas where language models traditionally struggle, such as symbolic manipulation, logical deduction, and algorithmic tracking.
\end{itemize}

\section{Additional Results and Analysis}

\subsection{The Impact of the Number of Turns on Performance}
We investigate the impact of the conversation length on model performance by increasing the interaction turns from 1 to 4 during the curriculum generation phase. As shown in Table~\ref{tab:turn_impact}, extending the number of turns yields significant benefits. Compared to the single-turn baseline, the 4-turn setting improves the executor's overall performance by 3.4\%, with specific gains of 3\% on mathematical benchmarks and 2.6\% on general domain tasks.
This performance boost can be attributed to the increased complexity of the curriculum. Multi-turn interactions encourage the curriculum agent to generate tasks with longer context dependencies and progressive difficulty. Consequently, the executor is forced to enhance its capability to maintain logical consistency and reasoning over extended horizons, rather than relying on simple pattern matching.

\begin{table}[t] 
\centering 
\footnotesize
\caption{Ablation study on the number of interaction turns. Increasing turns from 1 to 4 leads to consistent performance gains across all domains.} 
\label{tab:turn_impact} 
\begin{tabular}{c|ccc} 
\toprule 
\textbf{Number of Turns} & \textbf{Overall AVG} & \textbf{Math AVG} & \textbf{General AVG} \\
\midrule 
1 & 35.5 & 50.4 & 30.8 \\
2 & 35.8 & 50.7 & 31.1 \\
3 & 36.1 & 51.2 & 31.3 \\ 
4 & 36.7 & 51.9 & 31.6 \\
\bottomrule 
\end{tabular} 
\end{table}

\subsection{Detailed Results}
For a more detailed analysis, we report the complete results of the 3-iteration experiments in Table~\ref{tab:math_results_detail} and Table~\ref{tab:general_results_detail}. These tables provide a comprehensive breakdown of the agent's performance across all individual benchmarks, further validating the effectiveness and robustness of \ours.
\begin{table}[ht]
\centering
\small
\caption{Comprehensive results on mathematical reasoning benchmarks. The peak performance achieved during each model's training process is highlighted in \textbf{bold}.}
\label{tab:math_results_detail}
\resizebox{\textwidth}{!}{%
\begin{tabular}{lcccccccccc}
\toprule
\textbf{Model Name} & \adjustbox{valign=c, height=1.3ex, raise=0.4ex}{\includegraphics{images/tool.png}} & \adjustbox{valign=c, height=1.3ex, raise=0.4ex}{\includegraphics{images/openai.png}} & \textbf{AVG} & \textbf{AMC} & \textbf{Minerva} & \textbf{MATH} & \textbf{GSM8K} & \textbf{Olympiad} & \textbf{AIME25} & \textbf{AIME24} \\
\midrule
\textit{Qwen3-4B-Base} & & & & & & & & & & \\
Base Model & \textcolor{red}{\ding{55}} & \textcolor{teal}{\ding{55}}& 42.6 & 45.7 & 38.2 & 68.2 & 87.8 & 41.0 & 6.15 & 10.9 \\
+ $\texttt{\ours}$ (Iter 1) &\textcolor{teal}{\ding{51}} &\textcolor{teal}{\ding{55}} & 51.9 & 59.8 & 55.0 & 79.9 & 92.6 & 46.1 & 13.0 & 16.8 \\
+ $\texttt{\ours}$ (Iter 2) &\textcolor{teal}{\ding{51}} & \textcolor{teal}{\ding{55}}& {52.2} & 60.0 & 55.1 & 80.2 & 92.5 & {46.5} & {13.8} & {17.1} \\
+ $\texttt{\ours}$ (Iter 3) &\textcolor{teal}{\ding{51}} & \textcolor{teal}{\ding{55}}& {52.5} & {60.6} & {55.6} & {80.5} & {92.6} & {46.7} & {14.1} & {17.4} \\
\midrule
\textit{Qwen3-8B-Base} & & & & & & & & & & \\
Base Model & \textcolor{red}{\ding{55}}& \textcolor{teal}{\ding{55}}& 49.2 & 52.0 & 50.0 & 78.0 & 89.1 & 44.7 & 16.7 & 13.9\\
+ $\texttt{\ours}$ (Iter 1) & \textcolor{teal}{\ding{51}}& \textcolor{teal}{\ding{55}}& 55.1 & 57.3 & 59.0 & 81.6 & 93.9 & 48.4 & 20.9 & 24.9  \\ 
+ $\texttt{\ours}$ (Iter 2) &\textcolor{teal}{\ding{51}}& \textcolor{teal}{\ding{55}} & {56.5} & 59.2 & 60.1 & 81.9 & 94.0 & 51.2 & 22.9 & 26.1 \\
+ $\texttt{\ours}$ (Iter 3) &\textcolor{teal}{\ding{51}}& \textcolor{teal}{\ding{55}} & {58.2} & 62.4 & {61.3} & {82.4} & {94.5} & 54.0 & {24.8} & 28.0 \\
\bottomrule
\end{tabular}%
}
\vspace{-1em}
\end{table}

\begin{table}[h]
\centering
\small
\caption{Results on general-domain reasoning benchmarks. }
\label{tab:general_results_detail}
\resizebox{\textwidth}{!}{%
\begin{tabular}{lccccccc}
\toprule
\textbf{Model Name} & \adjustbox{valign=c, height=1.3ex, raise=0.4ex}{\includegraphics{images/tool.png}} & \adjustbox{valign=c, height=1.3ex, raise=0.4ex}{\includegraphics{images/openai.png}} & \textbf{Overall AVG} & \textbf{MATH AVG} & \textbf{SuperGPQA} & \textbf{MMLU-Pro} & \textbf{BBEH} \\
\midrule
\textit{Qwen3-4B-Base} & \\
Base Model & \textcolor{red}{\ding{55}} & \textcolor{teal}{\ding{55}}& 27.1 & 42.6 & 20.9 &	37.4 & 7.57 \\
+ $\texttt{\ours}$ (Iter 1) & \textcolor{teal}{\ding{51}} & \textcolor{teal}{\ding{55}}& 36.7 & 51.9 & 28.9 & 55.1 & 10.7 \\
+ $\texttt{\ours}$ (Iter 2) & \textcolor{teal}{\ding{51}} & \textcolor{teal}{\ding{55}}& {36.9} & {52.2} & {29.3} & {55.3} & {11.0} \\
+ $\texttt{\ours}$ (Iter 3) & \textcolor{teal}{\ding{51}} & \textcolor{teal}{\ding{55}} & {37.6} & {52.5} & {29.9} & {55.9} & {12.0} \\ 
\midrule
\textit{Qwen3-8B-Base} & \\
Base Model & \textcolor{red}{\ding{55}} & \textcolor{teal}{\ding{55}}& 34.5 & 49.2 & 28.3 & 51.8 & 8.6 \\
+ $\texttt{\ours}$ (Iter 1) & \textcolor{teal}{\ding{51}} & \textcolor{teal}{\ding{55}} & 40.7 & 55.1 & 32.5 & 62.2 & 13.0 \\
+ $\texttt{\ours}$ (Iter 2) & \textcolor{teal}{\ding{51}} & \textcolor{teal}{\ding{55}}& {41.3} & {56.5} & 32.5 & 62.4 & {13.6} \\
+ $\texttt{\ours}$ (Iter 3) & \textcolor{teal}{\ding{51}} & \textcolor{teal}{\ding{55}} & {42.1} & {58.2} & {33.0} & {63.4} & {13.7} \\
\bottomrule
\end{tabular}%
}
\vspace{-1em}
\end{table}

\section{Case Analysis}
To provide qualitative evidence of the model's evolution, Table~\ref{tab:case1}, Table~\ref{tab:case2}, Table~\ref{tab:case3}, Table~\ref{tab:case4}, Table~\ref{tab:case5}, Table~\ref{tab:case6}, Table~\ref{tab:case7}, Table~\ref{tab:case8}, and Table~\ref{tab:case9} through 9 present three representative questions generated at each stage from Iteration 1 to Iteration 3. We observe a clear progression in difficulty: while the initial iteration features relatively straightforward queries, the tasks in Iteration 3 evolve into highly complex, multi-step problems requiring deep reasoning. This escalation is driven by the co-evolutionary dynamic, where the Curriculum Agent, incentivized to maximize the Executor's learning signal, must continuously push the difficulty frontier to challenge the Executor's expanding proficiency, thereby effectively preventing learning stagnation.

\begin{table}[t]
\centering
\caption{Sampled questions generated by Curriculum Agent (Iter 1). }
\begin{tcolorbox}[colframe=black, colback=gray!10, coltitle=black, sharp corners]
\textbf{Questions from Curriculum Agent} \\
Let \( S \) be the set of all positive integers \( n \) for which the polynomial \\ \[ P(x) = x^3 - 2023x^2 + nx - 1 \] \\ has three distinct positive integer roots. Find the sum of all elements in \( S \).
\end{tcolorbox}
\label{tab:case1}
\end{table}

\begin{table}[t]
\centering
\caption{Sampled questions generated by Curriculum Agent (Iter 1). }
\begin{tcolorbox}[colframe=black, colback=gray!10, coltitle=black, sharp corners]
\textbf{Questions from Curriculum Agent} \\
In a triangle \(ABC\) with side lengths \(a\), \(b\), and \(c\) (where \(a = BC\), \(b = CA\), and \(c = AB\)), let the area be \(K\). If the incircle of the triangle touches \(BC\), \(CA\), and \(AB\) at \(D\), \(E\), and \(F\) respectively, and the lengths of \(BD\), \(CE\), and \(AF\) are \(x\), \(y\), and \(z\) respectively, prove that \(x^2 + y^2 + z^2 \geq \frac{3K}{2}\).
\end{tcolorbox}
\label{tab:case2}
\end{table}

\begin{table}[h]
\centering
\caption{Sampled questions generated by Curriculum Agent (Iter 1). }
\begin{tcolorbox}[colframe=black, colback=gray!10, coltitle=black, sharp corners]
\textbf{Questions from Curriculum Agent} \\
What is the minimum number of points inside a square with side length 1 that are needed to ensure that at least two of the points are at most 0.25 units apart from each other?
\end{tcolorbox}
\label{tab:case3}
\end{table}

\begin{table}[t]
\centering
\caption{Sampled questions generated by Curriculum Agent (Iter 2). }
\begin{tcolorbox}[colframe=black, colback=gray!10, coltitle=black, sharp corners]
\textbf{Questions from Curriculum Agent} \\
On a $9 \times 9$ chessboard, initially one cell is black. In each move, you can choose a white cell that has at least one black cell in the same row or column and invert the color of that chosen cell from white to black. Determine the minimum number of moves required to turn the entire chessboard into a black board.
\end{tcolorbox}
\label{tab:case4}
\end{table}

\begin{table}[t]
\centering
\caption{Sampled questions generated by Curriculum Agent (Iter 2). }
\begin{tcolorbox}[colframe=black, colback=gray!10, coltitle=black, sharp corners]
\textbf{Questions from Curriculum Agent} \\
Let \( S = \{1, 2, 3, \ldots, 100\} \). A subset \( A \) of \( S \) is called *good* if for any \( x, y \in A \) (with \( x \neq y \)), the sum \( x + y \) is not a perfect square. Find the maximum possible size of a *good* subset of \( S \).
\end{tcolorbox}
\label{tab:case5}
\end{table}

\begin{table}[t]
\centering
\caption{Sampled questions generated by Curriculum Agent (Iter 2). }
\begin{tcolorbox}[colframe=black, colback=gray!10, coltitle=black, sharp corners]
\textbf{Questions from Curriculum Agent} \\
In the land of Polytopia, each city is represented by a unique point on a large spherical map. The king decides to create a new city at a special point on this sphere. To determine the location, he uses a sequence of operations on the coordinates of existing cities. 
\\
The map is represented by a sphere where the equation $x^2 + y^2 + z^2 = 1$ holds for any point $(x, y, z)$ representing a city. The king chooses two existing cities, A and B, with coordinates $(a_1, a_2, a_3)$ and $(b_1, b_2, b_3)$, respectively. He defines a new city C with coordinates calculated by the formula:
\\
\[c_i = \frac{a_i^2 + b_i^2 + a_i b_i \cdot (1 + a_1 b_1 + a_2 b_2 + a_3 b_3)}{1 + a_1^2 + a_2^2 + a_3^2 + b_1^2 + b_2^2 + b_3^2 + (a_1 b_1 + a_2 b_2 + a_3 b_3)^2}\] for $i = 1, 2, 3$.
\\
Given that the coordinates of city A are $\left(\frac{1}{2}, \frac{1}{2}, \frac{\sqrt{2}}{2}\right)$ and the coordinates of city B are $\left(-\frac{1}{2}, \frac{1}{2}, \frac{\sqrt{2}}{2}\right)$, find the coordinates of the new city C.
\end{tcolorbox}
\label{tab:case6}
\end{table}

\begin{table}[t]
\centering
\caption{Sampled questions generated by Curriculum Agent (Iter 3). }
\begin{tcolorbox}[colframe=black, colback=gray!10, coltitle=black, sharp corners]
\textbf{Questions from Curriculum Agent} \\
A sequence of positive integers \( a_1, a_2, a_3, \ldots, a_{2024} \) is defined such that for each \( n \geq 1 \), the number \( a_{n+1} \) is determined by the rule \( a_{n+1} = a_n + \lfloor \sqrt{a_n} \rfloor \), starting with \( a_1 = 1 \). Find the remainder when \( a_{2024} \) is divided by 1000.
\end{tcolorbox}
\label{tab:case7}
\end{table}

\begin{table}[t]
\centering
\caption{Sampled questions generated by Curriculum Agent (Iter 3). }
\begin{tcolorbox}[colframe=black, colback=gray!10, coltitle=black, sharp corners]
\textbf{Questions from Curriculum Agent} \\
In a knockout tournament with $2^{n}$ players, where $n$ is a positive integer, each match eliminates one player. The tournament is structured such that each round halves the number of players. What is the minimum number of matches that must be played to determine the champion if, for each round, the number of matches played is a Fibonacci number?
\end{tcolorbox}
\label{tab:case8}
\end{table}

\begin{table}[t]
\centering
\caption{Sampled questions generated by Curriculum Agent (Iter 3). }
\begin{tcolorbox}[colframe=black, colback=gray!10, coltitle=black, sharp corners]
\textbf{Questions from Curriculum Agent} \\
A circle is divided into 2023 congruent arcs. The endpoints of one arc are colored red and blue. If two points that are diametrically opposite are both colored red, what is the probability that the two endpoints of the arc directly adjacent to the red endpoints are both blue? Express your answer as a common fraction.
\end{tcolorbox}
\label{tab:case9}
\end{table}

\section{Algorithm}

\begin{algorithm}[t]
\footnotesize
\caption{\textbf{Self-Evolutionary Framework \texttt{\ours}}}
\label{alg:ours_concise}
\begin{algorithmic}[1]
\REQUIRE Base LLM $\pi_{\text{base}}$; Iterations $T$; Samples $k$.
\STATE Initialize $\small \pi_{\theta}^{(0)} \leftarrow \pi_{\text{base}}$ and $\small \pi_{\phi}^{(0)} \leftarrow \pi_{\text{base}}$.
\FOR{each iteration $t = 1, \dots, T$}
    \STATE $\triangleright$ \textit{\textcolor{blue}{Curriculum Evolution (Train $\pi_{\theta}$)}}
    \STATE Initialize $\small \pi_{\theta} \leftarrow \pi_{\theta}^{(t-1)}$
    \STATE Generate a batch of tasks $X = \{x_i\} \sim \pi_{\theta}$
    \FOR{task $x_i \in X$}
        \STATE Sample $k$ responses $\{y_j\}_{j=1}^k \sim \pi_{\phi}^{(t-1)}(x_i)$
        \STATE Compute $R_{C}(x_i)$ using Eq.~\ref{eq:curr}
    \ENDFOR
    \STATE Update $\pi_\theta$ using $\mathcal{L}_{\text{GRPO}}$ with $(X, R_C)$ $\to$ $\pi_{\theta}^{(t)}$

    \STATE $\triangleright$ \textit{\textcolor{blue}{Executor Evolution (Train $\pi_{\phi}$) }}
    \STATE Generate $X_{\text{pool}} \sim \pi_{\theta}^{(t)}$ and filter to $\mathcal{D}^{(t)} = \{(x, \hat{p}, \tilde{y})\}$ where $|\hat{p}(x) - 0.5| \le \delta$
    
    \STATE Initialize $\pi_{\phi} \leftarrow \pi_{\phi}^{(t-1)}$
    \FOR{batch $B_\mathcal{D} = \{(x, \hat{p}(x), \tilde{y})\} \sim \mathcal{D}^{(t)}$}
        \STATE Initialize $\mathcal{T}_{\text{batch}}, \tilde{\mathcal{A}}_{\text{batch}}, \mathcal{P}_{\text{batch}}$
        \FOR{$(x, \hat{p}(x), \tilde{y}) \in B_\mathcal{D}$}
            \STATE Sample $k$ trajectories $\{\tau_i\}_{i=1}^k \sim \pi_{\phi}(x)$
            \STATE Compute rewards $R_i = \mathbb{I}(o_i = \tilde{y})$
            \STATE Compute scaled advantages $\tilde{A}_i \leftarrow A_i \cdot f(\hat{p}(x))$
            \STATE Add $\{\tau_i\}$ to $\mathcal{T}_{\text{batch}}$, $\{\tilde{A}_i\}$ to $\tilde{\mathcal{A}}_{\text{batch}}$, $\hat{p}(x)$ to $\mathcal{P}_{\text{batch}}$
        \ENDFOR
        \STATE Update $\pi_\phi$ using $\mathcal{L}_{\text{ADPO}}$ (Eq.~\ref{eq:adpo}) on collected batch
    \ENDFOR
    \STATE $\pi_{\phi}^{(t)} \leftarrow \pi_{\phi}$
\ENDFOR
\end{algorithmic}
\end{algorithm}

\end{document}